\documentclass[10pt,letterpaper]{article}
\usepackage{CJKutf8}
\usepackage[whole]{bxcjkjatype}
\usepackage[top=0.85in,left=2.75in,footskip=0.75in]{geometry}

\usepackage{amsmath,amssymb}

\usepackage{changepage}

\usepackage{textcomp,marvosym}

\usepackage{cite}

\usepackage{nameref,hyperref}

\usepackage[right]{lineno}

\usepackage[nopatch=eqnum]{microtype}
\DisableLigatures[f]{encoding = *, family = * }

\usepackage[table]{xcolor}

\usepackage{array}

\newcolumntype{+}{!{\vrule width 2pt}}

\newlength\savedwidth

\raggedright
\usepackage[aboveskip=1pt,labelfont=bf,labelsep=period,justification=raggedright,singlelinecheck=off]{caption}

\makeatletter
\renewcommand{\@biblabel}[1]{\quad#1.}
\makeatother

\usepackage{lastpage,fancyhdr,graphicx}
\usepackage{epstopdf}
\fancyheadoffset[L]{2.25in}
\fancyfootoffset[L]{2.25in}
\begin{document}
\vspace*{0.2in}

% Title must be 250 characters or less.--------------------------------------------------------------------
\begin{flushleft}
{\Large
\textbf\newline{ Metropolis-Hastings algorithm in joint-attention naming game: Experimental semiotics study } 
}
\newline
% Insert author names, affiliations and corresponding author email (do not include titles, positions, or degrees).
\\
Ryota Okumura\textsuperscript{1\Yinyang},
Tadahiro Taniguchi\textsuperscript{1*\Yinyang},
Yosinobu Hagiwara\textsuperscript{1},
Akira Taniguchi\textsuperscript{1}
\\
\bigskip
\textbf{1} Ritsumeikan University, 1-1-1 Noji-higashi, Kusatsu, Shiga, Japan
\\
\bigskip
\Yinyang These authors contributed equally to this work.
% Insert additional author notes using the symbols described below. Insert symbol callouts after author names as necessary.
% 
% Remove or comment out the author notes below if they aren't used.
%
% Primary Equal Contribution Note
%\Yinyang These authors contributed equally to this work.

% Additional Equal Contribution Note
% Also use this double-dagger symbol for special authorship notes, such as senior authorship.
%\ddag These authors also contributed equally to this work.

% Current address notes
%\textbf{1} Current Address: Ritsumeikan University, Kusatsu, Siga, Japan % change symbol to "\textcurrency a" if more than one current address note
% \textcurrency b Insert second current address 
% \textcurrency c Insert third current address

% Deceased author note
%\dag Deceased

% Group/Consortium Author Note
%\textpilcrow Membership list can be found in the Acknowledgments section.

% Use the asterisk to denote corresponding authorship and provide email address in note below.
* taniguchi@em.ci.ritsumei.ac.jp

\end{flushleft}
% Please keep the abstract below 300 words

\section*{Abstract}
In this study, we explore the emergence of symbols during interactions between individuals through an experimental semiotic study. Previous studies investigate how humans organize symbol systems through communication using artificially designed subjective experiments. In this study, we have focused on a joint attention-naming game (JA-NG) in which participants independently categorize objects and assign names while assuming their joint attention.

In the theory of the Metropolis-Hastings naming game (MHNG), listeners accept provided names according to the acceptance probability computed using the Metropolis-Hastings (MH) algorithm. The theory of MHNG suggests that symbols emerge as an approximate decentralized Bayesian inference of signs, which is represented as a shared prior variable if the conditions of MHNG are satisfied.

This study examines whether human participants exhibit behavior consistent with MHNG theory when playing JA-NG. By comparing human acceptance decisions of a partner's naming with acceptance probabilities computed in the MHNG, we tested whether human behavior is consistent with the MHNG theory.
The main contributions of this study are twofold. First, we reject the null hypothesis that humans make acceptance judgments with a constant probability, regardless of the acceptance probability calculated by the MH algorithm. This result suggests that people followed the acceptance probability computed by the MH algorithm to some extent. Second, the MH-based model predicted human acceptance/rejection behavior more accurately than the other four models: Constant, Numerator, Subtraction, and Binary. This result indicates that symbol emergence in JA-NG can be explained using MHNG and is considered an approximate decentralized Bayesian inference.

%\linenumbers

\section*{Introduction}\label{sec:introduction}

Humans have the ability to create and communicate through symbol systems that involve assigning meanings to signs. This semiotic process does not rely on predetermined definitions of the symbols' meanings but rather emerges gradually through semiotic communication and perceptual experiences. This phenomenon is known as symbol emergence~\cite{taniguchi2016symbol,taniguchi2018symbol}. Understanding the cognitive capabilities and the social and cognitive dynamics that support symbol emergence is crucial to comprehend the dynamic property of language.

Numerous experimental semiotic studies have been conducted to investigate how humans organize symbol systems through communication~\cite{galantucci2005experimental,scott2009signalling,healey2007graphical}. These studies demonstrated that humans can build communication systems from scratch~\cite{galantucci2005experimental,scott2009signalling,healey2007graphical,roberts2010experimental,quinn2001evolving}. Additionally, computational model-based studies in experimental semiotics, such as those by Kirby et al., Cornish et al., and Navarro et al.~\cite{kirby2008cumulative,cornish2010investigating,navarro2018extremists} validate the effectiveness of iterated learning models. Iterated learning is a process in which an individual acquires a behavior by observing a similar behavior in another individual who acquired it in the same way~\cite{kirby2008cumulative}. However, iterated learning is not an explanatory principle that answers the question of whether the emergence of a symbol system improves the environmental adaptation of a group of agents.
Iterated learning does not have a theoretical connection to explanatory theories about human perceptual systems.
In constant, symbol emergence based on the Metropolis-Hastings naming game (MHNG), which is the focus of this study, is closely related to predictive coding and the free-energy principle~\cite{hohwy2013predictive,friston2010free,friston2022world}, which are often referred to the general principle of cognition.
In this context, Taniguchi et al. hypothesized that symbol emergence could be viewed as a collective predictive coding by a group of agents~\cite{taniguchi2022emergent}.

Many studies have been conducted on computational models that represent symbol emergence systems. Pioneering studies have been conducted using naming games, in which remote robots share symbols to represent objects, and variants of referential games~\cite{Steels15,steels1999spontaneous,kirby2002natural,cangelosi1998emergence}. More recently, deep learning-based referential games have been extensively used to study emergent communication~\cite{havrylov2017emergence,lazaridou2017multiagent,evtimova2018emergent,bouchacourt2019miss}. Referential and naming games, often referred to as variants of the Lewis-style signaling game, have also been used to achieve compositionality in languages~\cite{kottur2017natural,Ren2020Compositional,choi2018multiagent,mu2021emergent}.
Generally, in these games, a speaker sends a message to a listener who indicates the object intended by the speaker. After the communication, reward feedback is provided to the agents, and they update their parameters. The reward feedback precedes joint attention in this approach. 

However, in the developmental process of human infants, joint attention, which is acquired at around nine months of age, is well known to precede tremendous progress in lexical acquisition and language development.  
Another notable idea is the naming game based on joint attention and the associated theoretical basis, called MHNG, in which each agent independently forms categories and shares signs associated with those categories through communication in the joint attention naming game (JA-NG)~\cite{hagiwara2019symbol}. This theory suggests that symbol emergence can be viewed as the approximate decentralized Bayesian inference of a posterior distribution over a shared latent variable conditioned on the observations of all agents participating in the communication. However, previous studies on experimental semiotics~\cite{kirby2008cumulative,cornish2010investigating,navarro2018extremists} have not employed computational models that incorporate decentralized Bayesian inference over the entire system, including multiple agents.

In this study, our objective is to investigate whether the MHNG, which models symbol emergence as a decentralized Bayesian inference~\cite{hagiwara2019symbol,taniguchi2022emergent}, can serve as a valid explanatory principle of symbol emergence between human individuals. The MHNG involves computational agents playing a JA-NG, where agents independently form categories of objects and name them while assuming joint attention. Unlike the widely used Lewis signaling games~\cite{lewis2008convention}, JA-NG does not involve any explicit reward feedback from the opponent after the naming process. In the MHNG, each agent decides whether to accept another agent's naming based on a probabilistic criterion calculated using the Metropolis-Hastings (MH) algorithm~\cite{hastings1970monte}. Consequently, symbol emergence occurs through a decentralized Bayesian inference.

Suppose people in JA-NG follow a similar acceptance probability as observed in MHNG. In this case, it can be inferred that they perform decentralized Bayesian inference as a whole system that includes multiple individuals involved in the emergence of symbols.
MHNG is a computational model in which agents play joint-attention naming games, and it uses the acceptance probability based on MH algorithm to determine whether a listener agent accepts an incoming name proposed by another agent. Testing the hypothesis that humans use MH-based criteria to determine the acceptance of new names in JA-NG is crucial for demonstrating the validity of the MHNG as an explanatory principle. If humans exhibit a behavior similar to that of the MHNG, their acceptance rate of incoming names should be correlated with the probability calculated using the MH algorithm. Thus, it can be concluded that humans make acceptance or rejection judgments in communication, following the principles of the MHNG to some extent. However, whether humans employ the same acceptance/rejection assessments in similar settings remains unclear.

In this study, we aim to verify whether humans engage in decentralized Bayesian inference by conducting subject experiments similar to JA-NG. To achieve this, we conducted a communication experiment with human participants. The communication structure in the experiment resembled that of the JA-NG in the simulation experiment conducted by Hagiwara et al~\cite{hagiwara2019symbol}. We observed the acceptance or rejection assessments of participants and tested whether they utilized the acceptance probability calculated by MHNG theory to a certain extent. Additionally, we evaluated whether the computational model using the MH algorithm predicted human behavior more accurately than four other comparative models, i.e., Constant, Numerator, Subtraction, and Binary.

The main contributions of this study are as follows:
\begin{itemize}
\item We verify whether human participants playing JA-NG utilize the acceptance probability computed in the model based on the MH algorithm to a certain extent.
\item We demonstrate that the model based on the MH algorithm outperforms the other four comparative computational models in predicting participants' acceptance behavior in JA-NG.
\end{itemize}

Statistical tests were conducted to examine our hypotheses. The results showed that the acceptance behavior of the human participants in JA-NG can be modeled using the MH algorithm.

The remainder of this paper is organized as follows: The next section provides an overview of the computational theory underlying this study. We then describe the setup of the communication experiment as well as the analysis and statistical test procedures in the Materials and Methods section. The Results and Discussion section presents our findings and corresponding interpretations. The final section concludes the paper.

\section*{Preliminaries}\label{sec:preliminaries}
In this section, we describe JA-NG performed in the subject experiments and the interpersonal Gaussian mixture (Inter-GM), which is the assumed probabilistic model for analyzing the results of the subject experiments. Additionally, we describe the general interpersonal probabilistic generative model (Inter-PGM), whose concrete instance is the inter-GM, and the MHNG in which agents play JA-NG using a specific acceptance probability based on the MH algorithms. 

\begin{figure}[tb]
\centering
\includegraphics[width=0.95\linewidth]{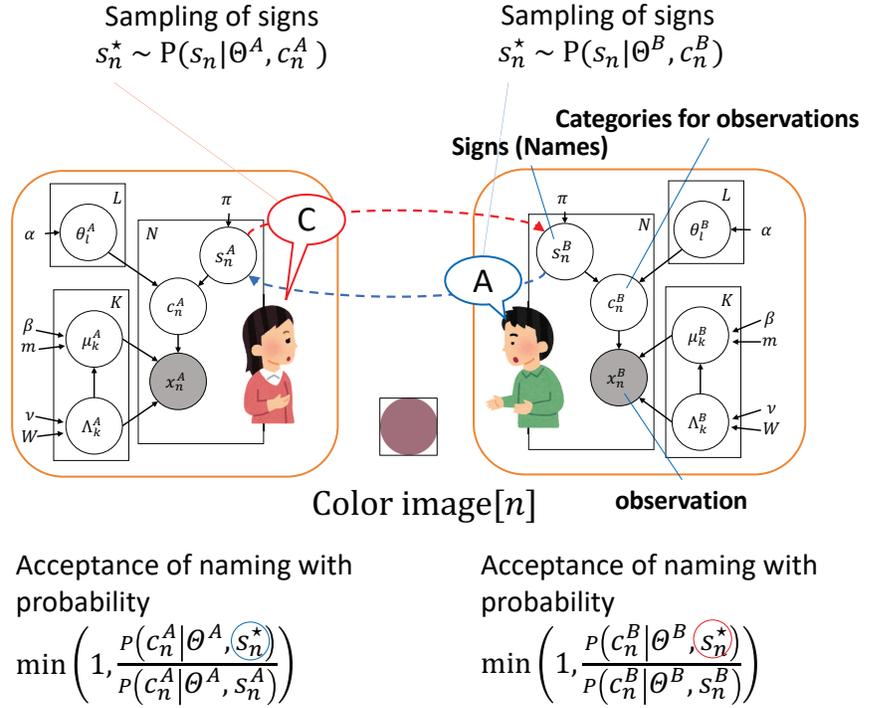}
    \caption{{\bf Illustration of the relationship between the communication game in the experiment and the probabilistic graphical model of the Inter-GM.} The color images observed by the participants are labeled as $x_n^A$ and $x_n^B$, with the corresponding color classification results represented by $c_n^A$ and $c_n^B$. The subjects' images are named by sampling a shared sign $s^\star_n$, with signs sampled for the $n$-th object by A and B, which are labeled as $s_n^A$ and $s_n^B$, respectively. The red balloon is A's sampled sign and the blue one is B's sampled sign. The transmission of the sign through naming is depicted by the dashed red and blue lines. $\Theta^*=\{\theta_l\}_{l=1, \ldots, L}$ and $\Phi^* = \{(\mu_k^*, \Lambda_k^*)\}_{k=1, \ldots, K}$}
    \label{img:overview}
\end{figure}

Fig.~\ref{img:overview} illustrates the correspondence between the computational model (i.e., inter-GM) and the communication experiment.

\subsection*{Joint-attention naming game (JA-NG)}\label{sec:ng}

Two agents $A$ and $B$ play JA-NG as detailed here. Specific variables are introduced in the following subsection.

\begin{enumerate}
    \item {\bf Perception}: Both the speaker and the listener observe an object and update their perceptual state, e.g., a categorization result, corresponding to the object based on their respective observations, assuming joint attention where two agents are looking at the same object.
    \item {\bf Communication}: The speaker gives the name to the object based on its perceptual state, e.g., the categorization result, and its own knowledge. The listener decides whether to accept the name.
    \item {\bf Learning}: After communication, the categorization results and knowledge are updated based on the results of the communication.
    \item {\bf Turn-taking}: The speaker and listener alternate their roles and repeat the above steps for all objects.
\end{enumerate}
The JA-NG is a procedural description of the interaction between two agents and their learning process through the sharing of semiotic knowledge between them based on joint attention.

\subsection*{Inter-PGM and MH naming game (MHNG)}\label{sec:mhng}

\begin{figure}[tb]
\centering
\includegraphics[width=0.95\linewidth]{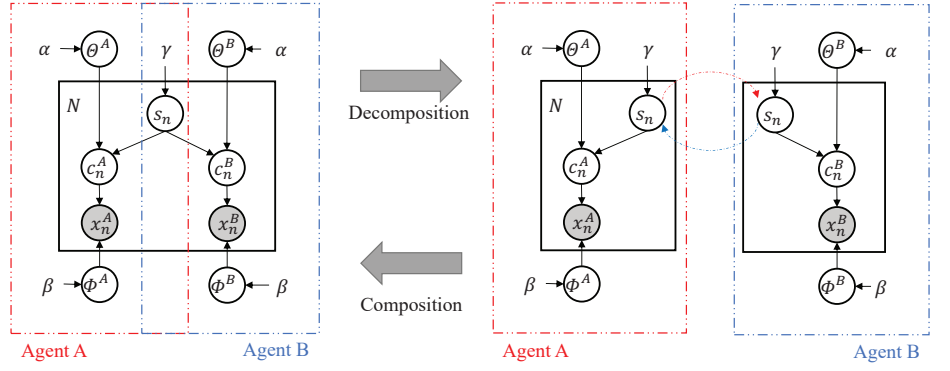}
    \caption{{\bf (Left) Probabilistic graphical model of inter-PGM. (Right) Decomposed illustration of the inter-PGM.}}
    \label{img:pgm}
\end{figure}

 We first define the variables related to JA-NG and assume a conditional dependency between the variables by defining the Inter-PGM (Fig.~\ref{img:pgm}).
 Table.\ref{tab:inter-pgm} is an explanation of the variables in the Inter-PGM.
 Inter-PGM is a general form of the PGMs that models the symbol emergence using JA-NG.

 \begin{table}[btp]
     \centering
     \begin{tabular}{cc} \hline
     Variable & Explanation  \\\hline
       $s_n$   &  A sign, e.g., a name, for the $n$-th object\\
       $c^*_n$   & Perceptual state corresponding to the $n$-th object \\
       $x^*_n$   & Observation for the $n$-th object   \\
       $\Theta^*$   & Parameter about the relations between signs and perceptual states  \\
       $\Phi^*$   & Parameter about the relations between perceptual states and observations \\
       $\alpha$   &  A hyperparameter for  $\Theta^*$ \\
       $\beta$   &   A hyperparameter for  $\Phi^*$       \\\hline
     \end{tabular}
     \caption{Variables of Inter-PGM and their explanations. Superscript $* \in \{A, B\}$ refers to a specific agent.}
     \label{tab:inter-pgm} 
 \end{table}

The probability variables related to JA-NG can be described using a probabilistic graphical model, as shown in Fig.~\ref{img:pgm}.

The generative process of the Inter-PGM is as follows:
\begin{align}
s_n &\sim P(s_n \mid \gamma) \quad  &n &= 1, \ldots, N\\
\Theta^* &\sim P(\Theta^* \mid \alpha)\\
\Phi^* &\sim P(\Phi^* \mid \beta )\\
c^*_n &\sim P(c^*_n \mid s_n , \Theta^*) \quad  &n &= 1, \ldots, N\\
x^*_n &\sim P(x^*_n \mid c_n^*, \Phi^*)\quad &n &= 1, \ldots, N
\end{align}
where $x^*_n$ represents the observed information, $c^*_n$ represents the category to which $x^*_n$ is classified, i.e., perceptual state, $s^*_n$ represents the sign of $x^*_n$, and $* \in \{A, B\}$.

The PGM can be decomposed into two parts corresponding to the two agents using the SERKET framework~\cite{taniguchi2020neuro} in the inference process. Hagiwara et al. found that a certain type of language game can be regarded as a decentralized inference process for an inter-PGM~\cite{hagiwara2019symbol}, and Taniguchi et al. formulated this idea as MHNG \cite{taniguchi2022emergent}.

The MH naming game is a special case of the JA-NG~\cite{taniguchi2022emergent}. 
JA-NG becomes the MHNG on satisfying the following conditions: 
\begin{enumerate}
    \item The speaker ($Sp$) selects the name $s_n^\star$ by sampling from the posterior distribution $P(s_n \mid \Theta^{Sp}, c_n^{Sp})$.
    \item The listener ($Li$) determines acceptance of sign $s_n^\star$ using the probability $r^{MH} = \min\left(1,\displaystyle\frac{P(c_n^{Li} \mid {\Theta^{Li}},s_n^{\star})}{P(c_n^{Li} \mid {\Theta^{Li}},s^{Li}_n)}\right)$.
    \item The agents update its internal variables $c^*_n, \Theta^*, \Phi^*$ using Bayesian inference appropriately.
\end{enumerate}

It is theoretically guaranteed that the MHNG is an approximate decentralized Bayesian inference of shared representations, i.e., $P(\{s_n\}_{n = 1, \ldots , N} \mid \{x_n^A, x_n^B\}_{n = 1, \ldots , N})$ and each agent's internal representations and knowledge. For more details, please refer to the original paper\cite{taniguchi2022emergent}.

\subsection*{Interpersonal Gaussian Mixture (Inter-GM)}\label{sec:inter-gm}
In this study, we used inter-GM, which was tailored to fit the observations, that is, the color information used in our experiment. Hagiwara et al. proposed inter-DM and inter-MDM  models in which agents observe bag-of-features representations, i.e, histograms~\cite{hagiwara2019symbol,hagiwara2022intermdm}. They formed individual categories using a Dirichlet mixture and shared signs linked to the formed categories through communication. Inter-GM is a modified version of inter-DM in which the part that formed categories using a Dirichlet mixture is replaced by a Gaussian mixture for categorizing multidimensional continuous real-valued vectors.

The inter-GM generative process is as follows:
\begin{align}
s_n &\sim {\rm Cat}(s_n \mid \pi) \quad  &n &= 1, \ldots, N\label{eq:inter-GMM+VAE-1}\\
\mu_k^*, \Lambda^*_k &\sim \mathcal{N}(\mu_k^* \mid m,(\beta\Lambda^*_k)^{-1}){\mathcal W}(\Lambda_k^* \mid \nu,W)\quad &k&=1, \ldots, K\\
\theta_l^* &\sim {\rm Dir}(\theta_l^* \mid \alpha)\quad &l &= 1, \ldots, L\\
c^*_n &\sim {\rm Cat}(c^*_n \mid \theta_{s_n}^*) \quad  &n &= 1, \ldots, N\\
x^*_n &\sim \mathcal{N}(x^*_n \mid \mu^*_{c_n},(\Lambda^*_{c_n})^{-1})\quad &n &= 1, \ldots, N\label{eq:inter-GMM+VAE-3}
\end{align}
${\rm Cat}(*)$ is the categorical distribution, $\mathcal{N}(*)$ is the Gaussian distribution, ${\mathcal W}(*)$ is the Wishart distribution, and ${\rm Dir}(*)$ is the Dirichlet distribution.
The parameters for Gaussian mixture model (GMM) $\{\mu_k^*, \Lambda^*_k\}_{k=1, \ldots, K}$ correspond to $\Phi^*$ and $\{ \theta_l^* \}_{l=1, \ldots, L}$ corresponds to $\Theta^*$ in Itner-PGM (Fig. \ref{img:pgm}) respectively.

In the MHNG, after observing (or sampling) $s^*_n$, the probabilistic variables for each agent become independent, and the parameters for each agent can be inferred using ordinal approximate Bayesian inference schemes. We applied Gibbs sampling, a widely used Markov chain Monte Carlo approximate Bayesian inference procedure~\cite{bishop2006pattern}, to sample the parameters $\mu_k^*$, $\Lambda^*_k$, $c^*_n$, and $\Theta^*$.  

In the MHNG, the sign $s_n$ is inferred by agents A and B through an alternative sampling of the sign $s_n$ from each other, and acceptance based on the acceptance probability of the MH algorithm $r_n^{MH} = \min\left(1,\displaystyle\frac{P(c^{Li}_n \mid {\Theta}^{Li},s_n^\star)}{P(c^{Li}_n \mid {\Theta^{Li}},s^{Li}_n)}\right)$ for the other agent's sign where ${\Theta}^{Li} = \{ \theta_l^{Li}\}_{l=1, \ldots, L}$ inferred using $c^{Li}_n$ and $s^\star_n$.

The acceptance probability estimated from the categorization results (see Fig.~\ref{img:human_furo}) and the actual acceptance/rejection decisions were recorded to investigate whether humans accept their opponents' proposals based on the MH acceptance probability.
The parameters ${\Theta^*}$ and ${\Phi^*}$ are inferred through Gibbs sampling using the categorization  $\{c_n^*\}_{n=1, \ldots, N}$ provided by the participants, along with their names $s_n^*$ and original observations $x_n^*$.
The MH acceptance probability $r^{MH}_n$ is then calculated, where $s_n^\star$ denotes the opponent's proposal.

\section*{Materials and Methods}

\subsection*{Communication experiment}

To investigate whether a listener's acceptance of the speaker's proposals aligns with the acceptance probability calculated by the MH algorithm $r^{MH}_n$, we conducted a communication experiment with human participants. Instead of the computational experiment described in~\cite{hagiwara2019symbol}, we conducted a  communication experiment with human participants that followed a methodology similar to that of experimental semiotics.

\begin{figure}[!tb]
  \begin{center}
    \includegraphics[width=0.9\linewidth]{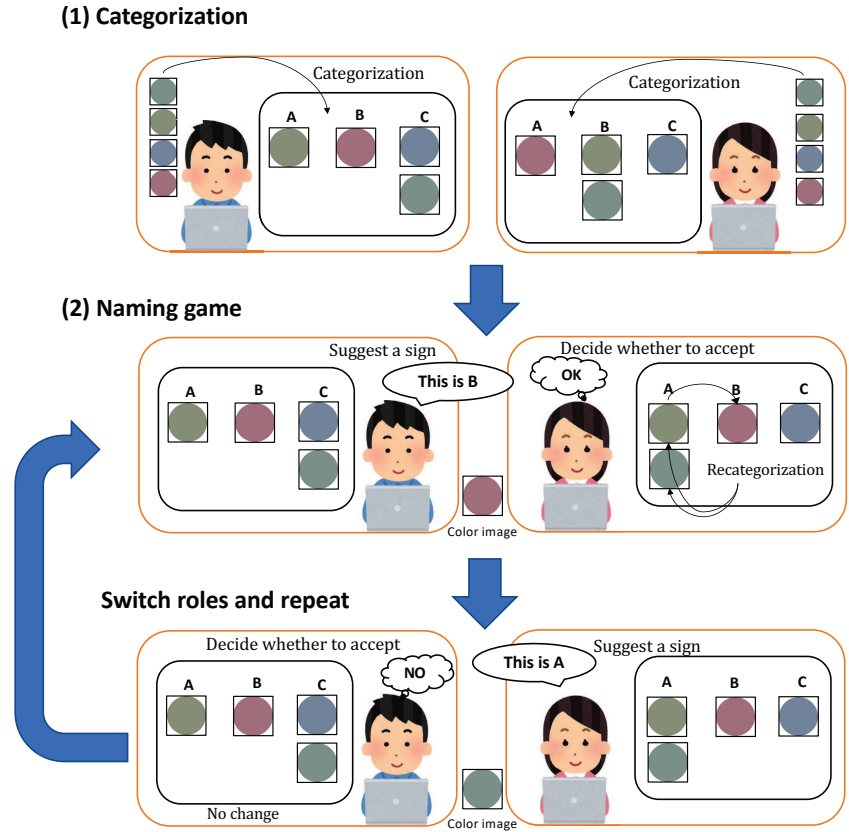}
    \caption{{\bf Flow of the subject experiment} In (1) categorization, participants categorize the given image. In (2) naming game, the speaker names the image by selecting any name from A to E, and the listener, decides whether to accept or reject the proposed name by pressing a button. Participants repeat the process, switching between the roles of speaker and listener.}
    \label{img:human_furo}
  \end{center}
\end{figure}

%\subsection*{Experimental method}
The experiment was conducted in pairs, with each pair comprising two participants, referred to as participants A and B. Each pair followed the procedure outlined in Fig.~\ref{img:human_furo} and used separate personal computers (PCs). Participants were in different rooms and were not permitted to communicate directly using any alternative communication media.

\begin{figure}[tb]
  \begin{center}
    \includegraphics[width=\linewidth]{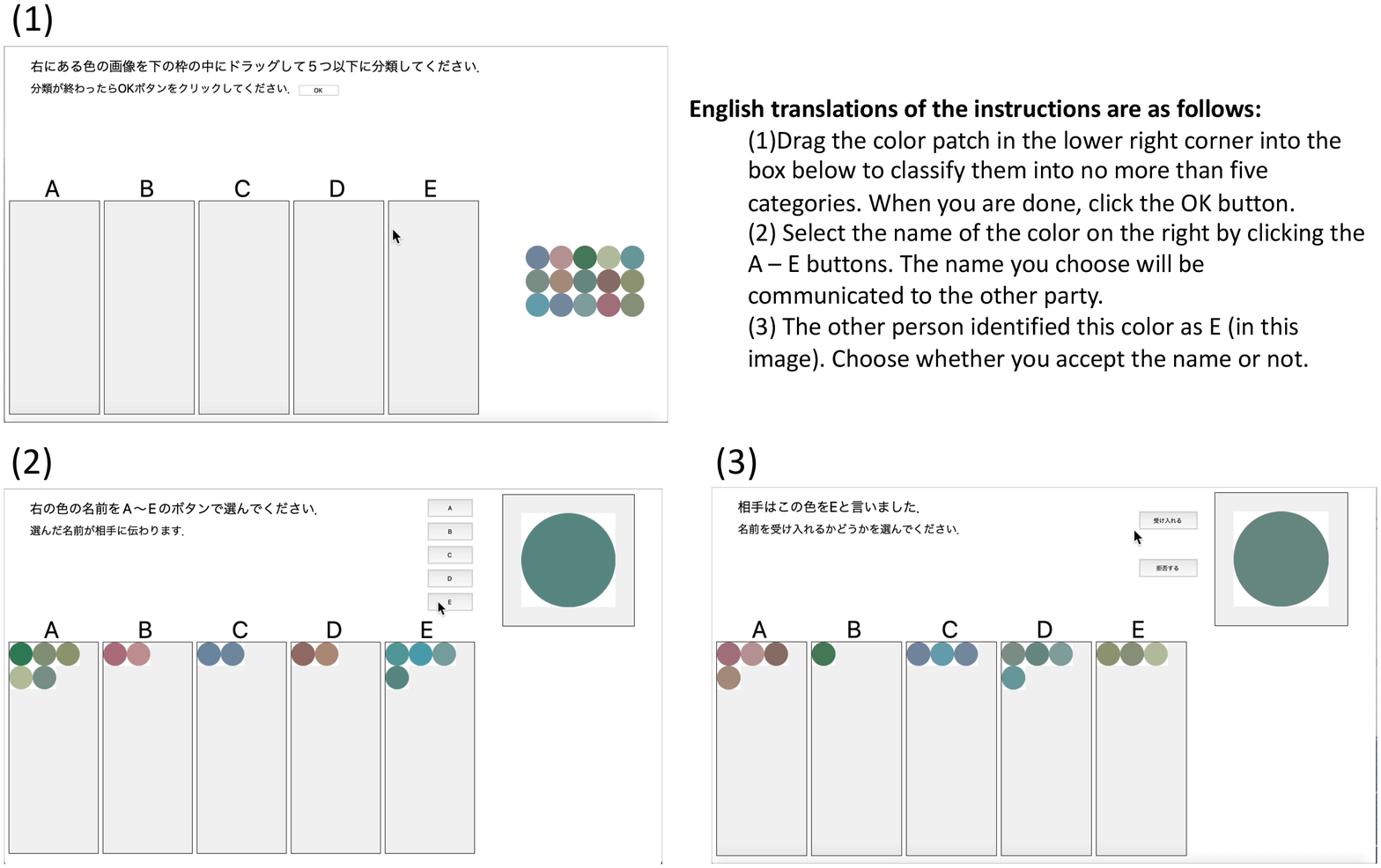}
    \caption{{\bf Screenshots of the experimental application in operation during the actual experiment.} (1) a view of the initial categorization phase, (2) a speaker's view in the naming phase, and (3) a listener's view when the listener receives a name of a color patch. 
}
    \label{img:app}
  \end{center}
\end{figure}

Fig.~\ref{img:app} shows the user interface of the experimental application. (1) in Fig.~\ref{img:app} shows the category classification screen that the participants first encountered, (2) shows the screen for the name, and (3) shows the screen for the listener. The procedure is detailed below.

Before starting the communication, each participant was instructed to classify the 15 images into categories labeled A–E (\textbf{initialization}).

\begin{enumerate}
\item \textbf{Perception}: An image used in the initialization step is displayed to a speaker. In the experiment, the participants were asked to exhibit their perceptual state as a categorization result ((1)~Categorization in Fig.~\ref{img:human_furo}).
\label{tejun1}
\item \textbf{Communication}:  The speaker names the image by selecting any name from A to E. Participant B, the listener, decides whether to accept or reject the proposed name by pressing a button. \label{tejun2}
\item \textbf{Learning} (update categories and sign allocation): Participant B, as the listener, can modify his/her classification result after the acceptance/rejection decision. \label{tejun3}
%\item \textbf{Turn-taking}: Participants switch roles and repeat steps from \ref{tejun1} to \ref{tejun3}. \label{tejun4}
\item \textbf{Turn-taking}: Steps from \ref{tejun1} to \ref{tejun3} correspond to (2)~Naming game in shown Fig.~\ref{img:human_furo}, and this game is repeated with participants switching their roles. \label{tejun4}

\end{enumerate}

During the experiment, the participants repeated steps \ref{tejun2} to \ref{tejun4} fifteen times for each data sample and then repeated the process three times. Therefore, each participant made 45 acceptance or rejection decisions per dataset.

\begin{figure}[tb]
  \begin{center}
    \includegraphics[width=0.9\linewidth]{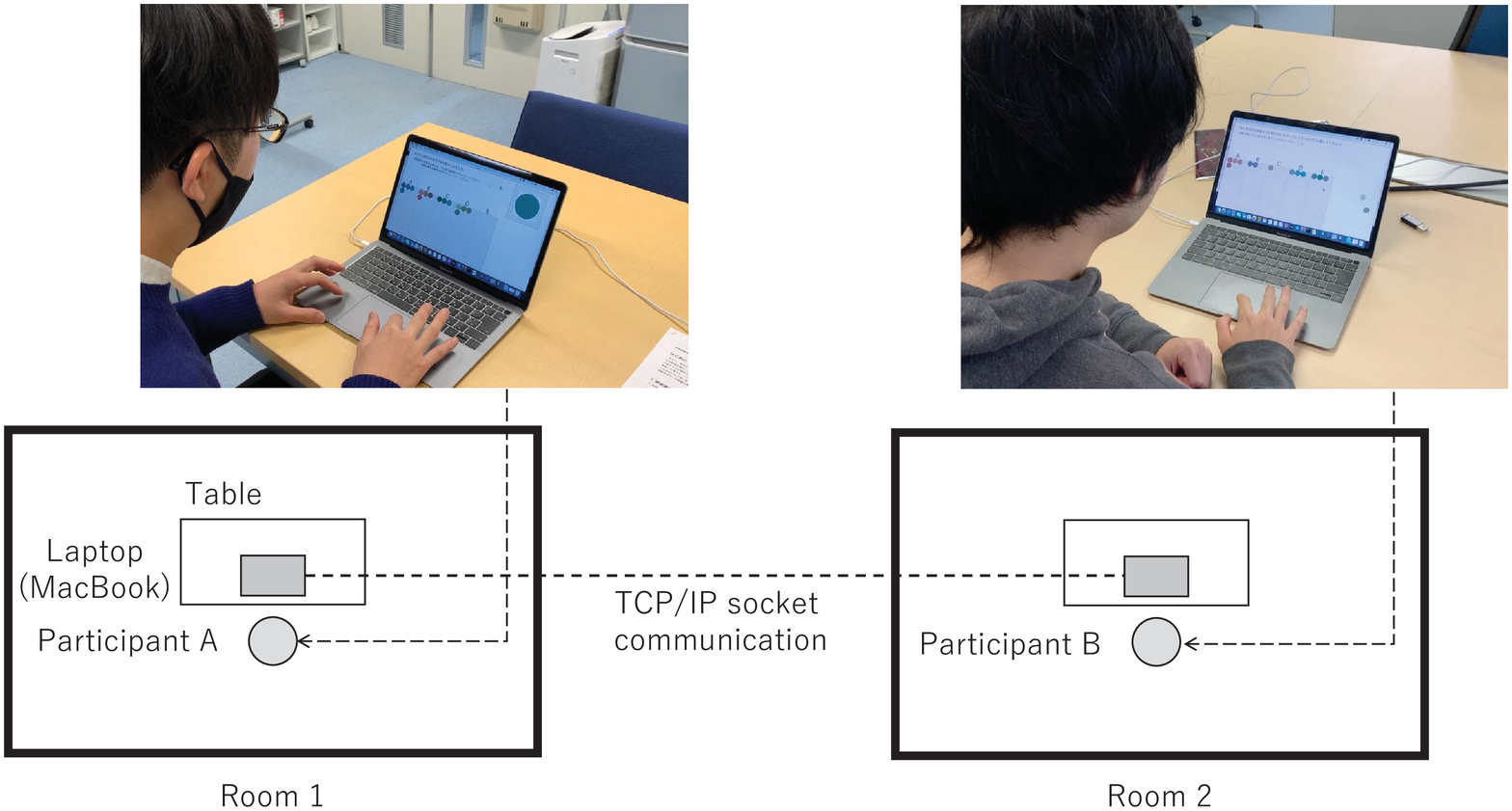}
    \caption{{\bf State of the actual experiment.} The two participants used different PCs and were housed in separate rooms, and they used socket communication for communication.}
    \label{img:state_of_experiment}
  \end{center}
\end{figure}

 The communication process involves proposing and accepting/rejecting names in steps \ref{tejun1} and \ref{tejun2}. Each communication was completed when step \ref{tejun2} ended and the results were recorded each time. Participants may modify their classification results whenever desired; however, a prompt appears if they attempt to alter the result after accepting/rejecting their partner's proposal when playing the listener's role. The two participants were housed in separate rooms, and the classification and communication were performed on PCs using a Python application that communicated with the other PCs. The PCs used was a 13-inch MacBook. The brightness of the PCs was automatically adjusted to account for the possibility of different ambient lighting in each room. The images were presented in random order because the same images were used even after switching roles in step \ref{tejun3}.
 Fig.~\ref{img:state_of_experiment} shows a photograph of an actual experiment.

\subsection*{Computational model for analysis}
We used the inter-GM described in the Preliminaries section to analyze the behavioral data and predict the acceptance rate of the participants. 

The hyperparameters used for the inter-GM were $\alpha=( 0.1, 0.1, 0.1, 0.1, 0.1)^{ \mathrm{T} }$, $\beta=1.0$, $m=(50, 0, 0)^{ \mathrm{T} }$, $W^{-1}=
\begin{pmatrix}
200 & 0 & 0 \\
0 & 200 & 0 \\
0 & 0 & 200\\
\end{pmatrix}
$, $\pi=( 1/5, 1/5, 1/5, 1/5, 1/5)^{ \mathrm{T} }$, in an empirical manner.

\subsection*{Materials}
For the experiment, 20 participants were recruited forming 10 pairs.
The female-to-male ratio was 6:14, and the minimum and maximum ages were 21 and 59 respectively.
As the experiment used colors, the participants were verbally asked if they were colorblind to ensure that colorblind participants were not included in the experiment.

This study was approved by the Research Ethics Committee of Ritsumeikan University under approval number BKC-LSMH-2022-012.
All the participants provided informed consent prior to participation.

To generate color images as stimuli, the CIE-$L^*U^*V^*$ color space, which accurately represents the psychological distance perceived by humans, was used \cite{steels2005coordinating}. In the CIE-$L^*U^*V^*$ color space, $L^*$ represents brightness and $U^*V^*$ represents hue. The details of the color images are as follows:
(1) Pillow (PIL), a Python image processing library, was used to create images of colored circles~\footnote{Pillow (PIL Fork) 8.4.0: \url{https://pillow.readthedocs.io/en/stable/}}.
(2)  $L^*$, $U^*$, and $V^*$ were sampled from three three-dimensional Gaussian distributions.
(3) Two datasets, {\bf hard} and {\bf easy}, were prepared to observe the differences in communication according to difficulty levels: Dataset 1 was difficult to classify, and Dataset 2 was easy to classify.
(4) The same images were shown to both participants and each dataset contained 15 images.
(5) The Gaussian distribution to sample from was determined using a uniform distribution.

\begin{figure}[tb]
    \includegraphics[width=\linewidth]{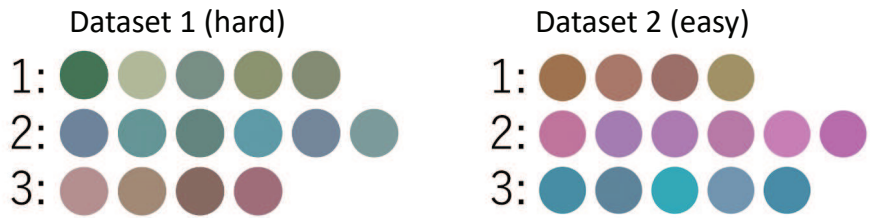}
    \caption{{\bf Images used in the communication experiment.} The left image is Dataset~1 (hard), and the right image is Dataset~2 (easy). The images in the same row are generated from the same Gaussian distribution, and the numbers represent the Gaussian distribution number.}
    \label{Dataset12_pdf}
\end{figure}

Table~\ref{Dataset_table} lists the parameters for each Gaussian distribution. Each data point in the three-dimensional CIE-$L^*U^*V^*$ color space was generated from a three-dimensional Gaussian distribution.
\begin{table}[bt]
\caption{Parameters of the three Gaussian distributions generating the color patches used in the experiment. $\mu_k$ is the mean vector of the $k$-th three-dimensional Gaussian distribution. $\Sigma = \Lambda^{-1}$ is the covariance matrix that is shared among the three Gaussian distributions}
\label{Dataset_table}
\begin{center}
\begin{tabular}{ccc}
\hline
\multicolumn{1}{c}{} & \multicolumn{1}{c}{Dataset~1 (hard)} & \multicolumn{1}{c}{Dataset~2 (easy)}\\
\hline\hline
\multicolumn{1}{c|}{$\mu_1$} & \multicolumn{1}{c}{$\begin{pmatrix}
60  \\
-10  \\
20 \\
\end{pmatrix}$} & \multicolumn{1}{c}{$\begin{pmatrix}
60  \\
30  \\
30 \\
\end{pmatrix}$}\\

\cline{1-3}

\multicolumn{1}{c|}{$\mu_2$} & \multicolumn{1}{c}{$\begin{pmatrix}
60  \\
-20  \\
-10 \\
\end{pmatrix}$} & \multicolumn{1}{c}{$\begin{pmatrix}
60  \\
30  \\
-30 \\
\end{pmatrix}$}\\

\cline{1-3}

\multicolumn{1}{c|}{$\mu_3$} & \multicolumn{1}{c}{$\begin{pmatrix}
60  \\
20  \\
10 \\
\end{pmatrix}$} & \multicolumn{1}{c}{$\begin{pmatrix}
60  \\
-30  \\
-30 \\
\end{pmatrix}$}\\

\cline{1-3}

\multicolumn{1}{c|}{$\Sigma$} & \multicolumn{1}{c}{$\begin{pmatrix}
5^2 & 0 & 0  \\
0 & 9^2 & 0  \\
0 & 0 & 9^2 \\
\end{pmatrix}$} & \multicolumn{1}{c}{$\begin{pmatrix}
5^2 & 0 & 0  \\
0 & 10^2 & 0  \\
0 & 0 & 10^2 \\
\end{pmatrix}$}\\
\hline
\end{tabular}
\end{center}
\end{table}
Fig~\ref{Dataset12_pdf} shows images of Dataset~1 ({\bf hard}), and Dataset~2 ({\bf  easy}).

\subsection*{Hypothesis testing 1}\label{sec:test1}
We investigated whether people's decisions are affected using the acceptance probability based on the MH algorithm, although the decision does not completely comply with the theory. 
To investigate whether humans use the MH-based acceptance probability to a certain extent, i.e., whether the actual acceptance probability correlates with the MH-based acceptance probability, we define a biased Bernoulli distribution, $ {\rm Bern}(z_n \mid r_n = a r^{MH}_n+b)$. 
The Bernoulli distribution, ${\rm Bern}(z \mid r)$, samples 1 with probability $r$ and 0 with probability $1-r$.
The  weight parameter $a$, indicating the extent to which the inferred acceptance probability is used, and bias parameter $b$, indicating the degree to which acceptance occurs unconditionally, were used and these parameters were estimated.
If $a=1$ and $b=0$, the distribution becomes the original MH-based acceptance probability distribution, $ {\rm Bern}(z_n \mid r_n = r^{MH}_n)$.
Specifically, variable $z_n$ represents whether the participant accepted the given name, taking the value of $1$ if accepted and $0$ if rejected. The acceptance probability of a participant is denoted by $r_n$.

We tested the estimated parameters $a$ and $b$, which model the relationship between the actual acceptance probability and MH-based acceptance probability $r^{MH}_n$.
For acceptance and rejection, we assumed $1$ and $0$, respectively. Instead of calculating the correlation between the acceptance/rejection decision and $r_n$, we used a conditional Bernoulli distribution.

Parameters $a$ and $b$ were determined using the maximum likelihood estimation. The maximum log-likelihood estimation of parameters $a$ and $b$ was performed using gradient descent.
The original likelihood function is defined as 
\begin{eqnarray*}
L(a,b)=\prod_{n=1}^{N}  {\rm Bern}(z_n \mid r_n = ar^{MH}_n+b)
\end{eqnarray*}
To avoid the {\rm Bern} parameter from going outside the domain, $a$ and $b$ were bounded to $0 \le  b$ and $a+b \le1$ , respectively.

A hypothesis test was performed to test the statistical significance of the association between the $r^{MH}_n$ score and acceptance decisions made by actual human participants.

The null hypothesis $H_0$ and alternative hypothesis $H_0$ are as follows:
\begin{itemize}
    \item $H_0$: There is no association between the acceptance decision and $r^{MH}_n$, the MH-based acceptance probability. In other words, the human acceptance probability remains constant with respect to $r^{MH}_n$.
    \item $H_1$: The acceptance probability is not constant, indicating that humans utilize the MH-based acceptance probability $r^{MH}_n$ to some extent $a \neq 0$.
\end{itemize}
The test statistic is the coefficient of a (bounded) linear function that parameterizes the Bernoulli distribution and the acceptance probability as output. The test statistic was set as the coefficient of the regression fitted to the observed data $\hat{a}$.

To estimate the sampling distribution of the test statistic, we used a randomized approach in which we randomly generated Bernoulli random variables with a fixed parameter and then fitted a linear model to obtain the coefficient $a$ (i.e., the test statistic) from the null hypothesis\footnote{This is because it is difficult to analytically determine the distribution that the estimated statistics $a$ and $b$ follow. We could not perform conventional tests such as the $t$-test.}. 
The acceptance and rejection decisions were randomly sampled from the distribution by assuming $H_0$, i.e., $z_n \sim {\rm Bern}(z \mid \bar{b})$. The null distribution of the test statistics was estimated and the $p$-values were empirically calculated.
By repeating this 1000 times, we obtained an estimate of the sampling distribution as a histogram, by which we could compute the $p$-value as the tail probability. $\bar{b}$ was determined from the behavior of all subjects using maximum likelihood estimation.

 By assuming that the acceptance event occurs with probability $r_n$, we can compute the likelihood by fitting them to the Bernoulli distribution and multiplying them by the total number of given names $N$; that is, $L=\prod_{n=1}^{N}  {\rm Bern}(z_n \mid r_n = ar^{MH}_n+b) $

We performed sampling using ${\rm Bern}(\bar{z} \mid \bar{b})$ to obtain lists of test statistics $a$ and $b$ and created their cumulative distribution functions to conduct a statistical test. The following steps describe the process of obtaining the list of test statistics $a$ and $b$:
From the experimental results, we calculated the acceptance rate $\bar{b}=\frac{1}{N}\sum_{n=1}^{N}z_n$ for all participants or target participants across all trials.
We sampled the acceptance or rejection of each round from the Bernoulli distribution ${\rm Bern}(\bar{z} \mid \bar{b})$ with the parameter $\bar{b}$ determined in the previous step, that is, $\bar{z}_n \sim {\rm Bern}(\bar{z} \mid \bar{b})\ (n \in {1, \ldots, N})$.
Parameters $a$ and $b$ were estimated using the maximum likelihood estimator for each sampling result and were added to the list of statistical quantities.
This procedure was repeated $1000$ times and the sample distributions of $a$ and $b$ were obtained.

We computed the cumulative distribution function $P_a'(a) = \frac{1}{L}\sum_{l=1}^{L}f(a_l,a)$ from a list of obtained statistical values $a$ represented as ${a_1, a_2, \ldots, a_L}$, where $L=1000$. Similarly, we compute the cumulative distribution function $P_b'(b) = \frac{1}{L}\sum_{l=1}^{L}f(b_l,b)$ from a list of statistical values $b$, represented by ${b_1,b_2, \ldots, b_L}$. Here,
\begin{eqnarray*}
f(x,y) = 
\begin{cases}
    1, & x \ge y \\
    0, & x < y
\end{cases}
\end{eqnarray*}
where $f(x,y)$ represents a function that returns 1 if $x$ exceeds or is equal to $y$, and 0 if $x$ is below $y$.

Because $r_{MH}$ can be used if it is significantly greater than $0$, a one-sided test was performed.
The bias parameter $b$ undergoes a two-sided test.
The significance level was set at $p < 0.001$.
Specifically, the following steps were performed. If $P_{a}'(\hat{a}) \le 0.001$, then the $p$-value $P_a(\hat{a}) < 0.001$, that is, the null hypothesis $H_0$ is rejected.
In addition, if $P_b'(\hat{b}) \ge 0.9995$ or $P_b'(\hat{b}) \le 0.0005$,  then the $p$-value $P_b(\hat{b}) < 0.0005$, that is, the null hypothesis $H_0$ is rejected.

\subsection*{Hypothesis testing 2}

In Test 2, we tested whether the model that used the MH algorithm, i.e., the acceptance decision using ${\rm Bern}(z_n \mid r^{MH}_n)$, was closer to the participants' behavior than several heuristic comparative models.
We performed a test using the assessment of acceptance or rejection obtained from the results of the communication experiment, and the inferred acceptance probability was denoted as $r^{MH}_n$. We created a set of data consisting of the distances between the participants' behaviors and the samples generated from the probabilities calculated by the five comparison models. These models were used to evaluate the acceptance and rejection. Subsequently, U-tests were conducted for each model.

\begin{table*}[!bt]
\caption{Details of each model}
\label{models}
\begin{center}
\begin{tabular}{cccc}
\hline
\multicolumn{1}{l}{\#} &\multicolumn{1}{l}{Model name} & \multicolumn{1}{l}{Acceptance probability formula} \\
\hline
\vspace{-3mm}\\
\multicolumn{1}{c}{1} &\multicolumn{1}{l}{\bf Constant} & \multicolumn{1}{l}{$r^1_n=\Bar{b}$} \\
\multicolumn{1}{c}{2} &\multicolumn{1}{l}{\bf MH} & \multicolumn{1}{l}{$r^2_n = r^{MH}_n$} \\
\multicolumn{1}{c}{3} &\multicolumn{1}{l}{\bf Numerator} & \multicolumn{1}{l}{$r^3_n=P(c_n^{Li} \mid \Theta^{Li},s_n^\star)$} \\
\multicolumn{1}{c}{4} &\multicolumn{1}{l}{\bf Subtraction} & \multicolumn{1}{l}{$r^4_n=(P(c_n^{Li} \mid \Theta^{Li},s_n^\star)-P(c_n^{Li} \mid \Theta^{Li},s_n^{Li}))/2+1/2$} \\
\multicolumn{1}{c}{5} &\multicolumn{1}{l}{\bf Binary} & \multicolumn{1}{l}{$r^5_n=\left\{\begin{array}{ll}
0.1 & (r \leq 0.5)\\
0.9 & (r > 0.5)
\end{array}\right. $} \\
\hline
\end{tabular}
\end{center}
\end{table*}

Table~\ref{models} lists the comparative models used in this study.
{\bf Constant} accepts with a probability $\Bar{b}$ calculated from the actual acceptance rate of the subject from the experimental results, which corresponds to the null hypothesis of hypothesis testing 1.
{\bf MH} accepts with the inferred MH-based acceptance probability $r^{MH}_n$ from the experimental results.
{\bf Numerator} accepts with the acceptance probability being the numerator part of the $r_n^{MH}$ score, which represents the likelihood of the opponent's sign using its own parameter.
{\bf Subtraction} calculates the difference between the numerator part of the $r^{MH}_n$ score representing the likelihood of the opponent's sign using the listener's parameter and the denominator part representing the likelihood of its own sign instead of the ratio in $r^{MH}_n$ score. Subsequently, it was transformed into a range of $0.0$–$1.0$.
{\bf Binary} accepts with a probability of $0.1$ if the inferred acceptance probability $r^{MH}_n$ is less than or equal to $0.5$ and $0.9$ if it exceeds $0.5$.

To test the statistical significance of models $m$ and $m'$ that make decisions regarding acceptance and rejection, hypothesis tests were performed as null and alternative hypotheses, respectively, as follows:

\begin{itemize}
\item $H_0$: ${\rm Prec}_{m} = {\rm Prec}_{m'}$. The models $m$ and $m'$ predict the participants' behavior at the same level. 
\item $H_1$: ${\rm Prec}_{m} > {\rm Prec}_{m'}$.  The model $m$ predicts the participant's behavior more accurately than the model $m'$.
\end{itemize}
Here, ${\rm Prec}_{m}$ is the rate at which the model $m$ could predict the participants' acceptance or rejection decisions, i.e., precision. 

We sampled $100$ data points for the pseudo-experimental results of each comparison model using computer simulations. The pseudo-experimental results for each comparison model were sampled from the Bernoulli distribution with the parameter of acceptance probability $r^m_{(j,i)}$ for subject $j$ of model $m$ in the $i$th communication trial and labeled $1$ for acceptance and $0$ for rejection.
The $p$values were calculated using a U-test. 
The significance level was set at $0.001$.

\begin{displaymath}
z^m_{(j,i)} \sim {\rm Bern}(z \mid r^m_{(j,i)})
\end{displaymath}

Precision was calculated as follows: First, we store the $j$-th participant's acceptance/rejection evaluation at the $i$th trial in the experiment in $z^h_{(j,i)}$, where $j = 1, \cdots, 20$. Second, we store the acceptance/rejection evaluation results of model $m$ in the $i$-th trial of the pseudo-experiment for subject $j$ in $z^m_{(j,i)}$, where $i = 1, \cdots, 45$ ($i = 1, \cdots, 90$ for both datasets). Third, we calculate the precision of model $m$ in predicting the $j$-th participant's behavior. 

The precision ${\rm Prec}_m$ is calculated by counting the number of matches between the participant's and model's decisions. One-sided tests were conducted for all model combinations.

\section*{Results and Discussion}
\subsection*{Hypothesis testing 1}
\begin{figure}[!tb]
\centering
\includegraphics[width=1.0\linewidth]{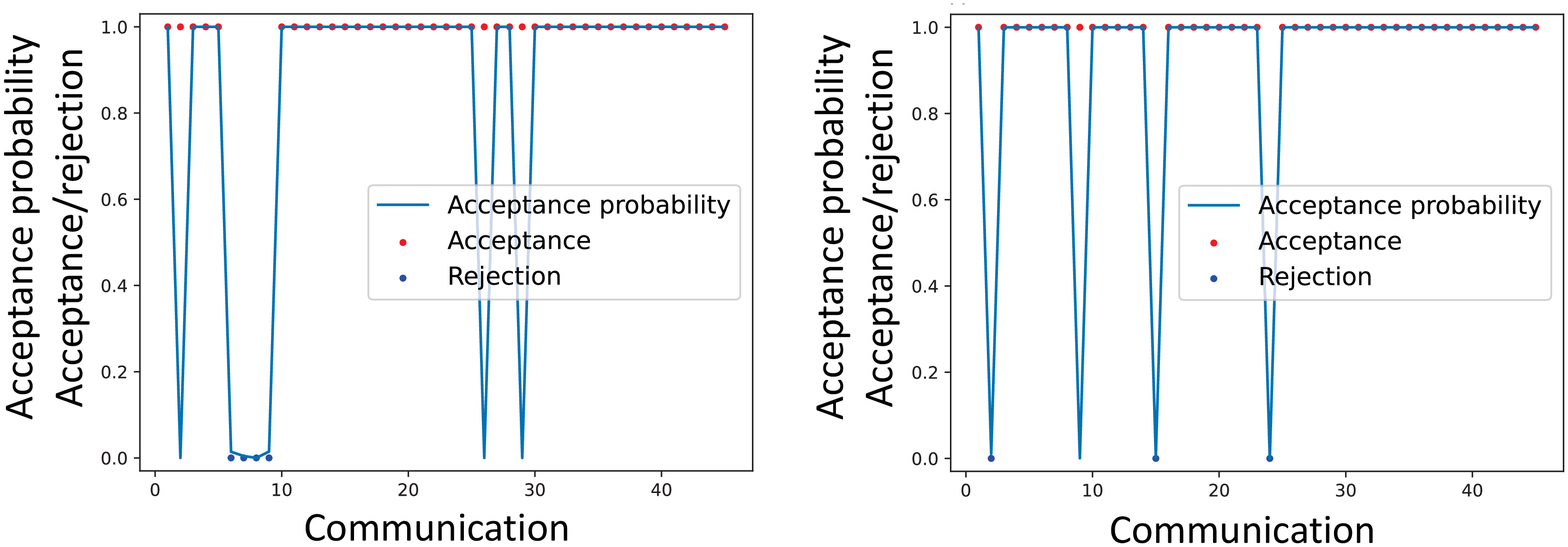}
    \caption{{\bf Example of the actual acceptance made by a participant and the inferred acceptance probabilities $r^{MH}_n$.} Dataset~1 (left), Dataset~2 (right).}    
    \label{example}
\end{figure}

Fig.~\ref{example} illustrates an example of the actual acceptance/rejection behavior of a participant and the inferred acceptance probabilities $r^{MH}$. This suggests that there is certain coherence between $r_n^{MH}$ and participants' behavior. 
This association was evaluated quantitatively and statistically.

\begin{figure}[tb]
\centering
\includegraphics[width=0.8\linewidth]{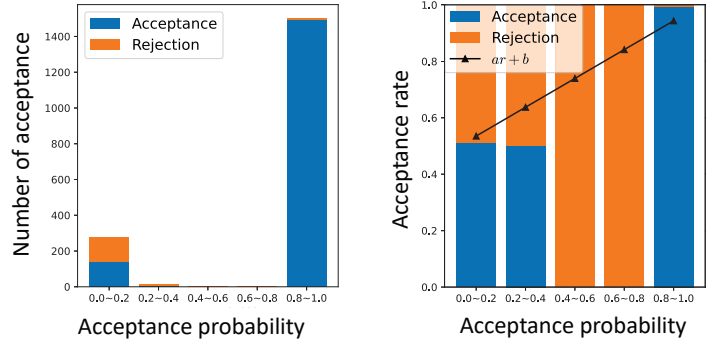}
    \caption{{\bf Relationship between the acceptance status of all participants and the inferred value of the acceptance probability} Graphs of the number of accepted names for each inferred acceptance probability for all participants (left), the actual acceptance rate for each inferred acceptance probability for all participants, and the graph of y = ar+b with weights `a' and bias `b' estimated by linear regression (right)}    
    \label{all_20}
\end{figure}

Fig.~\ref{all_20} shows a histogram of the number of accepted stimuli for each acceptance rate (left) and the actual acceptance rate for each acceptance rate with a graph of $y=ar+b$ using the estimated weights $a$ and bias $b$ (right) for all the participants, where $a = 0.5105$ and $b = 0.4842$.
When the inferred acceptance rate was high, the actual acceptance rate by humans were also high. However, the actual probability of acceptance was higher than $r^{MH}$ when $r^{MH}$ was low. 
It was rare for the inferred acceptance rate, $r_n^{MH}$ to assume an intermediate values between 0.2 and 0.8.

Subsequently, we describe the results of the hypotheses tests. First, we examine the results of Test 1. The estimated parameters for Datasets 1 and 2 are shown in Table~\ref{saiyu1}.
The $p$-values $P_a(a),P_b(b)$ for each subject obtained for each dataset in Table~\ref{saiyu1} show that they are rejected at the 0.001 significance level in all cases, except for some results for participants 6, 8 and 12.
$P_a(a)$ in Dataset 2 for Participant 8 and 12 is 0.009 and 0.008, which could not be rejected at the 0.001 significance level, but can be rejected at the 0.01 significance level.
The tests for both datasets and all subjects were rejected at a significance level of 0.001.
Therefore, the null hypothesis is rejected, suggesting that humans use the inferred acceptance probabilities $r^{MH}$ to a certain extent.

\begin{table}[bt]
\caption{Parameters $a,b$ estimated and $p$-values on each subject's data and the data aggregated over all participants for each dataset}
\label{saiyu1}
\begin{center}
\begin{tabular}{c|cccccccc}
\hline
\multicolumn{1}{c|}{ } & \multicolumn{4}{c|}{Dataset~1 (hard)} & \multicolumn{4}{c}{Dataset~2 (easy)}\\
\hline
\multicolumn{1}{c|}{ } & \multicolumn{1}{c}{$a$} & \multicolumn{1}{c}{$b$}& \multicolumn{1}{c}{$P_a(a)$} & \multicolumn{1}{c|}{$P_b(b)$}& \multicolumn{1}{c}{$a$} & \multicolumn{1}{c}{$b$}& \multicolumn{1}{c}{$P_a(a)$} & \multicolumn{1}{c}{$P_b(b)$}\\
\hline
\multicolumn{1}{c|}{1} & \multicolumn{1}{c}{$0.6249$} & \multicolumn{1}{c}{$0.3749$}& \multicolumn{1}{c}{$<0.001$} & \multicolumn{1}{c|}{ $<0.0005$}& \multicolumn{1}{c}{$0.9999$} & \multicolumn{1}{c}{ $0.0000$}& \multicolumn{1}{c}{$<0.001$} & \multicolumn{1}{c}{ $<0.0005$}\\
\hline
\multicolumn{1}{c|}{2} & \multicolumn{1}{c}{$0.2166$} & \multicolumn{1}{c}{ $0.7832$}& \multicolumn{1}{c}{$<0.001$} & \multicolumn{1}{c|}{ $<0.0005$}& \multicolumn{1}{c}{$0.2059$} & \multicolumn{1}{c}{ $0.7938$}& \multicolumn{1}{c}{$<0.001$} & \multicolumn{1}{c}{ $<0.0005$}\\
\hline
\multicolumn{1}{c|}{3} & \multicolumn{1}{c}{$0.3914$} & \multicolumn{1}{c}{ $0.6085$}& \multicolumn{1}{c}{$<0.001$} & \multicolumn{1}{c|}{ $<0.0005$}& \multicolumn{1}{c}{$0.2775$} & \multicolumn{1}{c}{ $0.7223$}& \multicolumn{1}{c}{$<0.001$} & \multicolumn{1}{c}{ $<0.0005$}\\
\hline
\multicolumn{1}{c|}{4} & \multicolumn{1}{c}{$0.3442$} & \multicolumn{1}{c}{ $0.6557$}& \multicolumn{1}{c}{$<0.001$} & \multicolumn{1}{c|}{ $<0.0005$}& \multicolumn{1}{c}{$0.7499$} & \multicolumn{1}{c}{ $0.2499$}& \multicolumn{1}{c}{$<0.001$} & \multicolumn{1}{c}{ $<0.0005$}\\
\hline
\multicolumn{1}{c|}{5} & \multicolumn{1}{c}{$0.2227$} & \multicolumn{1}{c}{ $0.7771$}& \multicolumn{1}{c}{$<0.001$} & \multicolumn{1}{c|}{ $<0.0005$}& \multicolumn{1}{c}{$-0.0002$} & \multicolumn{1}{c}{ $1.0000$}& \multicolumn{1}{c}{$<0.001$} & \multicolumn{1}{c}{ $<0.0005$}\\
\hline
\multicolumn{1}{c|}{6} & \multicolumn{1}{c}{$0.1216$} & \multicolumn{1}{c}{ $0.8783$}& \multicolumn{1}{c}{$<0.001$} & \multicolumn{1}{c|}{ $0.01$}& \multicolumn{1}{c}{$-0.0003$} & \multicolumn{1}{c}{ $1.0002$}& \multicolumn{1}{c}{$<0.001$} & \multicolumn{1}{c}{ $<0.0005$}\\
\hline
\multicolumn{1}{c|}{7} & \multicolumn{1}{c}{$0.5713$} & \multicolumn{1}{c}{ $0.4285$}& \multicolumn{1}{c}{$<0.001$} & \multicolumn{1}{c|}{ $<0.0005$}& \multicolumn{1}{c}{$0.7501$} & \multicolumn{1}{c}{ $0.2496$}& \multicolumn{1}{c}{$<0.001$} & \multicolumn{1}{c}{ $<0.0005$}\\
\hline
\multicolumn{1}{c|}{8} & \multicolumn{1}{c}{$0.4169$} & \multicolumn{1}{c}{ $0.5552$}& \multicolumn{1}{c}{$<0.001$} & \multicolumn{1}{c|}{ $<0.0005$}& \multicolumn{1}{c}{$-0.0245$} & \multicolumn{1}{c}{ $1.0007$}& \multicolumn{1}{c}{$0.009$} & \multicolumn{1}{c}{ $<0.0005$}\\
\hline
\multicolumn{1}{c|}{9} & \multicolumn{1}{c}{$0.4219$} & \multicolumn{1}{c}{ $0.5779$}& \multicolumn{1}{c}{$<0.001$} & \multicolumn{1}{c|}{ $<0.0005$}& \multicolumn{1}{c}{$-0.0001$} & \multicolumn{1}{c}{ $1.0001$}& \multicolumn{1}{c}{$<0.001$} & \multicolumn{1}{c}{ $<0.0005$}\\
\hline
\multicolumn{1}{c|}{10} & \multicolumn{1}{c}{$-0.0003$} & \multicolumn{1}{c}{ $1.0002$}& \multicolumn{1}{c}{$<0.001$} & \multicolumn{1}{c|}{ $<0.0005$}& \multicolumn{1}{c}{$0.5000$} & \multicolumn{1}{c}{ $0.5000$}& \multicolumn{1}{c}{$<0.001$} & \multicolumn{1}{c}{ $<0.0005$}\\
\hline
\multicolumn{1}{c|}{11} & \multicolumn{1}{c}{$0.2551$} & \multicolumn{1}{c}{ $0.7446$}& \multicolumn{1}{c}{$<0.001$} & \multicolumn{1}{c|}{ $<0.0005$}& \multicolumn{1}{c}{$0.4827$} & \multicolumn{1}{c}{ $0.4927$}& \multicolumn{1}{c}{$<0.001$} & \multicolumn{1}{c}{ $<0.0005$}\\
\hline
\multicolumn{1}{c|}{12} & \multicolumn{1}{c}{$0.3427$} & \multicolumn{1}{c}{ $0.6572$}& \multicolumn{1}{c}{$<0.001$} & \multicolumn{1}{c|}{ $<0.0005$}& \multicolumn{1}{c}{$-0.0072$} & \multicolumn{1}{c}{ $1.0071$}& \multicolumn{1}{c}{$0.008$} & \multicolumn{1}{c}{ $<0.0005$}\\
\hline
\multicolumn{1}{c|}{13} & \multicolumn{1}{c}{$0.7151$} & \multicolumn{1}{c}{ $0.2846$}& \multicolumn{1}{c}{$<0.001$} & \multicolumn{1}{c|}{ $<0.0005$}& \multicolumn{1}{c}{$0.4231$} & \multicolumn{1}{c}{ $0.5766$}& \multicolumn{1}{c}{$<0.001$} & \multicolumn{1}{c}{ $<0.0005$}\\
\hline
\multicolumn{1}{c|}{14} & \multicolumn{1}{c}{$0.2316$} & \multicolumn{1}{c}{ $0.7682$}& \multicolumn{1}{c}{$<0.001$} & \multicolumn{1}{c|}{ $<0.0005$}& \multicolumn{1}{c}{$-0.0004$} & \multicolumn{1}{c}{ $1.0004$}& \multicolumn{1}{c}{$<0.001$} & \multicolumn{1}{c}{ $<0.0005$}\\
\hline
\multicolumn{1}{c|}{15} & \multicolumn{1}{c}{$0.5998$} & \multicolumn{1}{c}{ $0.3999$}& \multicolumn{1}{c}{$<0.001$} & \multicolumn{1}{c|}{ $<0.0005$}& \multicolumn{1}{c}{$0.9999$} & \multicolumn{1}{c}{ $0.0000$}& \multicolumn{1}{c}{$<0.001$} & \multicolumn{1}{c}{ $<0.0005$}\\
\hline
\multicolumn{1}{c|}{16} & \multicolumn{1}{c}{$0.3414$} & \multicolumn{1}{c}{ $0.6585$}& \multicolumn{1}{c}{$<0.001$} & \multicolumn{1}{c|}{ $<0.0005$}& \multicolumn{1}{c}{$-0.0002$} & \multicolumn{1}{c}{ $1.0001$}& \multicolumn{1}{c}{$<0.001$} & \multicolumn{1}{c}{ $<0.0005$}\\
\hline
\multicolumn{1}{c|}{17} & \multicolumn{1}{c}{$0.8615$} & \multicolumn{1}{c}{ $0.1383$}& \multicolumn{1}{c}{$<0.001$} & \multicolumn{1}{c|}{ $<0.0005$}& \multicolumn{1}{c}{$0.5000$} & \multicolumn{1}{c}{ $0.5000$}& \multicolumn{1}{c}{$<0.001$} & \multicolumn{1}{c}{ $<0.0005$}\\
\hline
\multicolumn{1}{c|}{18} & \multicolumn{1}{c}{$-0.0002$} & \multicolumn{1}{c}{ $1.0001$}& \multicolumn{1}{c}{$<0.001$} & \multicolumn{1}{c|}{ $<0.0005$}& \multicolumn{1}{c}{$0.5000$} & \multicolumn{1}{c}{ $0.5000$}& \multicolumn{1}{c}{$<0.001$} & \multicolumn{1}{c}{ $<0.0005$}\\
\hline
\multicolumn{1}{c|}{19} & \multicolumn{1}{c}{$0.8642$} & \multicolumn{1}{c}{ $0.1356$}& \multicolumn{1}{c}{$<0.001$} & \multicolumn{1}{c|}{ $<0.0005$}& \multicolumn{1}{c}{$0.8970$} & \multicolumn{1}{c}{ $0.0757$}& \multicolumn{1}{c}{$<0.001$} & \multicolumn{1}{c}{ $<0.0005$}\\
\hline
\multicolumn{1}{c|}{20} & \multicolumn{1}{c}{$0.4478$} & \multicolumn{1}{c}{ $0.2798$}& \multicolumn{1}{c}{$<0.001$} & \multicolumn{1}{c|}{ $<0.0005$}& \multicolumn{1}{c}{$0.7932$} & \multicolumn{1}{c}{ $0.0974$}& \multicolumn{1}{c}{$<0.001$} & \multicolumn{1}{c}{ $<0.0005$}\\
\hline
\multicolumn{1}{c|}{All} & \multicolumn{1}{c}{$0.4910$} & \multicolumn{1}{c}{ $0.5030$}& \multicolumn{1}{c}{$<0.001$} & \multicolumn{1}{c|}{ $<0.0005$}& \multicolumn{1}{c}{$0.5478$} & \multicolumn{1}{c}{ $0.4476$}& \multicolumn{1}{c}{$<0.001$} & \multicolumn{1}{c}{ $<0.0005$}\\
\hline
\end{tabular}

\vspace{3mm}

\begin{tabular}{c|cccc}
\hline
\multicolumn{1}{c|}{ } & \multicolumn{4}{c}{ALL Dataset}\\
\hline
\multicolumn{1}{c|}{ } & \multicolumn{1}{c}{$a$} & \multicolumn{1}{c}{$b$}& \multicolumn{1}{c}{$P_a(a)$} & \multicolumn{1}{c}{$P_b(b)$}\\
\hline
\multicolumn{1}{c|}{All} & \multicolumn{1}{c}{$0.5105$} & \multicolumn{1}{c}{ $0.4842$}& \multicolumn{1}{c}{$<0.001$} & \multicolumn{1}{c}{ $<0.0005$}\\
\hline
\end{tabular}
\end{center}
\end{table}

\subsection*{Hypothesis testing 2}

Subsequently, we examine the results of Test 2. Table~\ref{U-test} shows the $p$-values obtained from the U-tests conducted for each combination of models. The row for {\bf MH} (i.e., $m=2$) in Table~\ref{U-test} shows that the null hypothesis is rejected for all the models. The model using the MH algorithm was the closest to the participants' behaviors among the models compared in this study.
We also individually performed tests on data from each participant. Table~\ref{U-test-each-count} presents the results. 
For each participant, {bf MH} outperformed the other models in predicting behavior in all cases, except for six participants in {bf Constant} and one in {bf Subtraction}. For the six participants, {\bf MH} did not significantly outperform {bf Constant}, and for one participant, {bf MH} did not significantly outperform {bf Subtraction}.
We tested the data for each participant separately, and even for each dataset.  Tables~\ref{U-test-each-count-dataset1} and~\ref{U-test-each-count-dataset2} list the results.
Looking at the {\bf MH} (i.e., $m=2$) row in Table~\ref{U-test-each-count-dataset1}, {\bf MH} outperforms the other models in all cases except 7 for {bf Constant} and 1 for {\bf Numerator}.
Looking at the {\bf MH} (i.e., $m=2$) row in Table~\ref{U-test-each-count-dataset2}, {\bf MH} outperformed the other models in all cases, except five for {\bf Constant}.
Based on these test results, we suggest that humans use the acceptance probability $r^{MH}$ derived from the MH algorithm during communication.

The experimental results supported our hypothesis that human behavior in JA-NG follows the MH algorithm. 
Consequently, this result suggests that symbol emergence through JA-NG between people performs decentralized Bayesian inference, i.e., collective predictive coding.

\begin{table}[bt]
\caption{P-value for U-test for each model combination for all participants}
\label{U-test}
\begin{center}
\begin{tabular}{c|ccccc}
\hline
\multicolumn{1}{c|}{$m\backslash m'$} & \multicolumn{1}{c}{Constant} & \multicolumn{1}{c}{MH} & \multicolumn{1}{c}{Numerator} & \multicolumn{1}{c}{Subtraction} & \multicolumn{1}{c}{Binary}\\
\hline
\hline
\multicolumn{1}{c|}{Constant} & \multicolumn{1}{c}{---} & \multicolumn{1}{c}{ $1.000$} & \multicolumn{1}{c}{ $<0.001$} & \multicolumn{1}{c}{ $<0.001$} & \multicolumn{1}{c}{ $<0.001$}\\
\hline
\multicolumn{1}{c|}{MH} & \multicolumn{1}{c}{$<0.001$} & \multicolumn{1}{c}{ ---} & \multicolumn{1}{c}{ $<0.001$} & \multicolumn{1}{c}{$<0.001$} & \multicolumn{1}{c}{ $<0.001$}\\
\hline
\multicolumn{1}{c|}{Numerator} & \multicolumn{1}{c}{$1.000$} & \multicolumn{1}{c}{ $1.000$} & \multicolumn{1}{c}{ ---} & \multicolumn{1}{c}{$<0.001$} & \multicolumn{1}{c}{ $0.003$}\\
\hline
\multicolumn{1}{c|}{Subtraction} & \multicolumn{1}{c}{$1.000$} & \multicolumn{1}{c}{ $1.000$} & \multicolumn{1}{c}{ $1.000$} & \multicolumn{1}{c}{ ---} & \multicolumn{1}{c}{$1.000$}\\
\hline
\multicolumn{1}{c|}{Binary} & \multicolumn{1}{c}{$1.000$} & \multicolumn{1}{c}{$1.000$} & \multicolumn{1}{c}{$0.997$} & \multicolumn{1}{c}{ $<0.001$} & \multicolumn{1}{c}{ ---}\\
\hline
\end{tabular}
\end{center}

\caption{Number of participants whose behavior resulted in the rejection of the null hypothesis for each pair of models}
\label{U-test-each-count}
\begin{center}
\begin{tabular}{c|ccccc}
\hline
\multicolumn{1}{c|}{$m\backslash m'$} & \multicolumn{1}{c}{Constant} & \multicolumn{1}{c}{MH} & \multicolumn{1}{c}{Numerator} & \multicolumn{1}{c}{Subtraction} & \multicolumn{1}{c}{Binary}\\
\hline
\hline
\multicolumn{1}{c|}{Constant} & \multicolumn{1}{c}{---} & \multicolumn{1}{c}{ $4$} & \multicolumn{1}{c}{ $12$} & \multicolumn{1}{c}{ $18$} & \multicolumn{1}{c}{ $12$}\\
\hline
\multicolumn{1}{c|}{MH} & \multicolumn{1}{c}{$14$} & \multicolumn{1}{c}{ ---} & \multicolumn{1}{c}{ $20$} & \multicolumn{1}{c}{$19$} & \multicolumn{1}{c}{ $20$}\\
\hline
\multicolumn{1}{c|}{Numerator} & \multicolumn{1}{c}{$6$} & \multicolumn{1}{c}{ $0$} & \multicolumn{1}{c}{ ---} & \multicolumn{1}{c}{$20$} & \multicolumn{1}{c}{ $7$}\\
\hline
\multicolumn{1}{c|}{Subtraction} & \multicolumn{1}{c}{$2$} & \multicolumn{1}{c}{ $0$} & \multicolumn{1}{c}{ $0$} & \multicolumn{1}{c}{ ---} & \multicolumn{1}{c}{$0$}\\
\hline
\multicolumn{1}{c|}{Binary} & \multicolumn{1}{c}{$6$} & \multicolumn{1}{c}{$0$} & \multicolumn{1}{c}{$1$} & \multicolumn{1}{c}{ $20$} & \multicolumn{1}{c}{ ---}\\
\hline
\end{tabular}
\end{center}

\caption{Number of participants whose behavior resulted in the rejection of the null hypothesis for each pair of models in Dataset~1}
\label{U-test-each-count-dataset1}
\begin{center}
\begin{tabular}{c|ccccc}
\hline
\multicolumn{1}{c|}{$m\backslash m'$} & \multicolumn{1}{c}{Constant} & \multicolumn{1}{c}{MH} & \multicolumn{1}{c}{Numerator} & \multicolumn{1}{c}{Subtraction} & \multicolumn{1}{c}{Binary}\\
\hline
\hline
\multicolumn{1}{c|}{Constant} & \multicolumn{1}{c}{---} & \multicolumn{1}{c}{ $6$} & \multicolumn{1}{c}{ $12$} & \multicolumn{1}{c}{ $18$} & \multicolumn{1}{c}{ $12$}\\
\hline
\multicolumn{1}{c|}{MH} & \multicolumn{1}{c}{$13$} & \multicolumn{1}{c}{ ---} & \multicolumn{1}{c}{ $19$} & \multicolumn{1}{c}{$20$} & \multicolumn{1}{c}{ $20$}\\
\hline
\multicolumn{1}{c|}{Numerator} & \multicolumn{1}{c}{$6$} & \multicolumn{1}{c}{ $0$} & \multicolumn{1}{c}{ ---} & \multicolumn{1}{c}{$20$} & \multicolumn{1}{c}{ $8$}\\
\hline
\multicolumn{1}{c|}{Subtraction} & \multicolumn{1}{c}{$2$} & \multicolumn{1}{c}{ $0$} & \multicolumn{1}{c}{ $0$} & \multicolumn{1}{c}{ ---} & \multicolumn{1}{c}{$0$}\\
\hline
\multicolumn{1}{c|}{Binary} & \multicolumn{1}{c}{$5$} & \multicolumn{1}{c}{$0$} & \multicolumn{1}{c}{$4$} & \multicolumn{1}{c}{ $20$} & \multicolumn{1}{c}{ ---}\\
\hline
\end{tabular}
\end{center}

\caption{Number of participants whose behavior resulted in the rejection of the null hypothesis for each pair of models in Dataset~2}
\label{U-test-each-count-dataset2}
\begin{center}
\begin{tabular}{c|ccccc}
\hline
\multicolumn{1}{c|}{$m\backslash m'$} & \multicolumn{1}{c}{Constant} & \multicolumn{1}{c}{MH} & \multicolumn{1}{c}{Numerator} & \multicolumn{1}{c}{Subtraction} & \multicolumn{1}{c}{Binary}\\
\hline
\hline
\multicolumn{1}{c|}{Constant} & \multicolumn{1}{c}{---} & \multicolumn{1}{c}{ $3$} & \multicolumn{1}{c}{ $10$} & \multicolumn{1}{c}{ $18$} & \multicolumn{1}{c}{ $9$}\\
\hline
\multicolumn{1}{c|}{MH} & \multicolumn{1}{c}{$15$} & \multicolumn{1}{c}{ ---} & \multicolumn{1}{c}{ $20$} & \multicolumn{1}{c}{$20$} & \multicolumn{1}{c}{ $20$}\\
\hline
\multicolumn{1}{c|}{Numerator} & \multicolumn{1}{c}{$6$} & \multicolumn{1}{c}{ $0$} & \multicolumn{1}{c}{ ---} & \multicolumn{1}{c}{$20$} & \multicolumn{1}{c}{ $9$}\\
\hline
\multicolumn{1}{c|}{Subtraction} & \multicolumn{1}{c}{$2$} & \multicolumn{1}{c}{ $0$} & \multicolumn{1}{c}{ $0$} & \multicolumn{1}{c}{ ---} & \multicolumn{1}{c}{$0$}\\
\hline
\multicolumn{1}{c|}{Binary} & \multicolumn{1}{c}{$4$} & \multicolumn{1}{c}{$0$} & \multicolumn{1}{c}{$4$} & \multicolumn{1}{c}{ $20$} & \multicolumn{1}{c}{ ---}\\
\hline
\end{tabular}
\end{center}
\end{table}

\section*{Conclusion and Discussion}

In this study, we conducted a communication experiment on symbol emergence, in which participants played a JA-NG in pairs. We compared the acceptance decisions of human participants with those of the computational models 
and confirmed that the acceptance probability of the model based on the MH algorithm was used to a certain extent by the participants. Additionally, the MH-based model outperformed the other five comparative computational models in terms of predicting the participants' behavior through two statistical tests.
Consequently, the model using the MH algorithm was found to be suitable for explaining human acceptance behavior in JA-NG. 

This suggests that the MHNG, which was studied computationally as a constructive approach to human symbol emergence, is a reasonable model for explaining symbol emergence in computational agents and human groups. This finding also supports the collective predictive coding hypothesis, which argues that symbol emergence in human society can be regarded as a decentralized Bayesian inference of a prior variable shared among people~\cite{taniguchi2022emergent}.  
 To advance our understanding of the human acceptance evaluation in JA-NG and the dynamics of symbol emergence among people, future studies should aim to gather more evidence by conducting experiments in diverse scenarios to test whether they follow the MH algorithm.

Exploring symbol emergence in a human-agent mixed system is a future challenge worth pursuing. Because we obtained evidence supporting the prediction of human participants' behavior using the MH algorithm, we could approximate human behavior as a computational agent following the MH algorithm. Based on this approximation, we can theoretically model and analyze a mixed system involving a human participant and a computer agent. 

\section*{Acknowledgments}
This work was supported by JSPS KAKENHI Grant Numbers JP21H04904 and JP17H06379.
The authors thank Prof. Takeuchi of Nagoya University for advice on hypothesis testing methods.

\nolinenumbers

\bibliography{plos}

\end{document}